\documentclass[pmlr]{jmlr}

\usepackage{longtable}
\usepackage{comment}

\usepackage{amsmath}
\usepackage{amssymb}
\usepackage{graphicx}
\usepackage{booktabs}
\usepackage{listings}
\usepackage{xcolor}
\usepackage{hyperref}
\usepackage{multirow}
\usepackage{graphicx}
\usepackage{subcaption}
\usepackage{wrapfig}
\usepackage{float}
\usepackage{longtable}

\usepackage{booktabs}
\usepackage[load-configurations=version-1]{siunitx} 
\usepackage{comment}

\newcommand{\cs}[1]{\texttt{\char`\\#1}}

\theorembodyfont{\upshape}
\theoremheaderfont{\scshape}
\theorempostheader{:}
\theoremsep{\newline}
\newtheorem*{note}{Note}

\jmlrvolume{1}
\jmlryear{2026}
\jmlrworkshop{Conference on Neurosymbolic Learning and Reasoning}

\title[How Compliant is Sepsis Treatment?] {How Compliant is Sepsis Treatment? An Expert-Guided Neuro-symbolic Pipeline for Generating Clinical Compliance Insights}

\author{
\Name{Himanshu Tripathi} \Email{htripathi@crimson.ua.edu}\\
\Name{Kaushik Roy} \Email{kroy2@ua.edu}\\
\Name{Subash Neupane} \Email{sneupane4@ua.edu}\\
\Name{Shahram Rahimi} \Email{srahimi1@ua.edu}\\
\addr{Department of Computer Science, The University of Alabama}
}

\begin{document}

\maketitle

\begin{abstract}
Verifying whether clinical care follows evidence-based protocols is a natural neuro-symbolic problem, yet the safety-critical setting defeats either paradigm alone. We present an expert-guided pipeline that constrains a large language model strictly to semantic normalization, mapping messy drug and microbiology strings onto a fixed clinical vocabulary, while a Sugeno fuzzy inference system reasons over the normalized events. The fuzzy layer encodes eight Surviving Sepsis Campaign bundle rules and replaces binary judgments with graded scores in $[0,1]$. Applied to 2,438 MIMIC-IV v3.1 sepsis episodes, it surfaces antibiotic timing as the most critical breakdown (mean 0.24, 13\% within one hour), Hour-1 underperformance (mean 36.7\%), a 51\% elevated-lactate drop-off, and descriptive differences in ICU stay across compliance groups (3.8 versus 5.1 days).
\end{abstract}

\textbf{Code: }\url{https://tinyurl.com/yy2zvpd9}


\section{Introduction} \label{sec:intro}

\begin{figure}[!htb]
    \centering
    \includegraphics[width=0.85\linewidth]{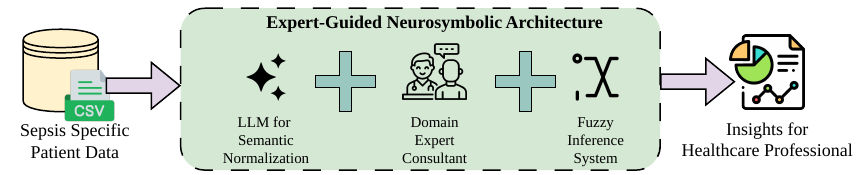}
    \caption{\footnotesize The end-to-end workflow of the Expert-Guided Neurosymbolic pipeline, illustrating the transformation of sepsis specific raw clinical data into healthcare insights via LLM semantic normalization, expert knowledge, and fuzzy logic.}
    \label{fig:abstract}
\end{figure}

\noindent Sepsis remains a leading cause of in-hospital mortality, and timely adherence to international, evidence‑based clinical guideline such as Surviving Sepsis Campaign (SSC) bundle \citep{SSC} is critical for patient survival. Large-scale Electronic Health dataset such as $MIMIC-IV$ \citep{MM4} offer a significant opportunity to evaluate protocol compliance at scale across thousands of patients. However, clinical data in these records is deeply unstructured and inconsistent. Traditional rule-based systems fail on semantic variations like trade names and synonyms, while pure neural networks produce uninterpretable risk scores that cannot guarantee adherence to safety protocols, leaving a critical gap in compliance assessment. To bridge this gap, we present an expert-guided neuro-symbolic pipeline (Figure \ref{fig:abstract}) where each component addresses a specific limitation of the existing paradigms. MedGemma acts as a semantic normalizer that safely resolves messy clinical text, and an expert-validated Fuzzy Inference System then evaluates the normalized data to produce graded compliance scores rather than brittle binary judgments. The system encodes eight SSC bundle rules spanning three clinical phases: immediate actions including blood culture sequencing, antibiotic administration, and lactate measurement within the first hour, followed by hemodynamic resuscitation rules covering fluid administration and vasopressor initiation, and finally treatment response rules assessing Mean Arterial Pressure (MAP) recovery and lactate clearance. For the complete rule definitions, refer to the \hyperref[app:rules]{Appendix: Eight Rules From SSC}.


\noindent This work is guided by 4 Research Questions (RQ): 


\noindent \textbf{RQ1:} \label{rq1} Is a neuro-symbolic approach necessary where purely symbolic or neural systems fall short? (Answered in \hyperref[subsec:regvsgem]{Regex vs MedGemma Validation})

\noindent \textbf{RQ2:} \label{rq2} Can a hybrid classifier pipeline achieve reliable semantic normalization of clinical data? (Answered in \hyperref[subsec:validation]{Validation})

\noindent \textbf{RQ3:} \label{rq3} Can a fuzzy inference system generate graded SSC compliance scores that provide actionable clinical insights aligns with real-world clinical decision-making patterns and reflects subject matter expert (SME) perspectives on sepsis management? (Answered in \hyperref[subsec:rulecomp]{Rule Compliance})

\noindent \textbf{RQ4:} \label{rq4} What compliance patterns emerge across the sepsis cohort, and how do they correlate with patient outcomes? (Answered in \hyperref[subsec:icu]{ICU Outcomes})


\section{Literature Survey} \label{sec:litsur}

\begin{figure}[!htb]
    \centering
    \includegraphics[width=0.75\linewidth]{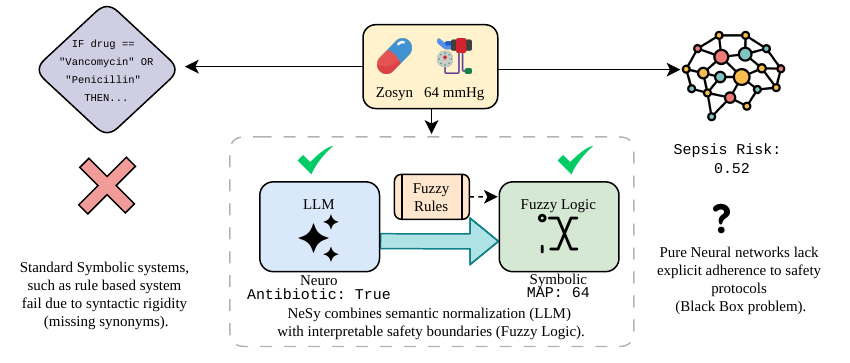}
    \caption{\footnotesize Comparison of sepsis detection paradigms handling complex inputs like "Zosyn" and 64 mmHg blood pressure. Traditional symbolic systems fail from vocabulary brittleness , while pure neural networks fail as uninterpretable black boxes. The proposed Neuro-Symbolic approach succeeds by pairing LLM semantic normalization with deterministic fuzzy logic safety boundaries.}
    \label{fig:introfig}
\end{figure}

\noindent Sepsis remains a leading cause of ICU mortality, where timely adherence to evidence-based protocols is critical for patient survival \citep{evans2021surviving}. The Surviving Sepsis Campaign bundles represent the gold standard for reducing mortality \citep{you2022relationship}, yet compliance continues to be suboptimal across healthcare settings \citep{damiani2015effect}, making its assessment increasingly important \citep{gao2005impact}. Electronic Health Records such as MIMIC-IV offer unprecedented opportunities for evaluating protocol adherence \citep{gupta2022extensive}, yet their inherent heterogeneity creates substantial challenges. Purely data-driven approaches such as Deep Learning achieve high predictive accuracy but  operate as black boxes \citep{yang2024development}, lacking the interpretability required for high-stakes clinical decision-making \citep{horng2017creating}. Conversely, traditional Rule-Based Systems offer transparency but are brittle when facing real-world clinical noise \citep{nelson2011normalized}, failing on trade name variations such as Vancocin versus Vancomycin. A hybrid system robust to noisy data yet interpretable regarding safety rules is therefore necessary \citep{tripathi2023experimental}. NeSy has emerged as such a paradigm for healthcare \citep{bhuyan2024neuro}, combining neural perception with symbolic reasoning to enforce logical constraints and domain knowledge \citep{alshahrani2017neuro, wu2023k, delong2023neurosymbolic}. In sepsis management, neural models can process time-series vitals while symbolic modules verify SSC bundle compliance \citep{evans2021surviving}. However, most existing NeSy frameworks rely on LSTMs or RNNs \citep{pabon2022negation} that struggle with the semantic nuance of unstructured clinical text \citep{chen2019attention}, particularly negation, uncertainty, and assertion detection in free-text notes \citep{ji2024assertion}. Large language models address this semantic gap \citep{maity2025large, wang2023clinicalgpt} but introduce hallucination risk in clinical environments \citep{omar2025multi, asgari2025framework}, making autonomous LLM deployment unsafe for critical interventions such as antibiotic  administration. Hybrid architectures that constrain LLMs to semantic normalization while delegating decisions to symbolic engines \citep{mcinerney2023chill, hasan2025clin} directly address this limitation. Fuzzy logic provides the complementary mechanism for uncertainty-aware decision support \citep{arji2019fuzzy}, positioning the LLM as a semantic normalizer and the fuzzy engine as the safety-guaranteeing reasoner (the architectural pattern illustrated in Figure \ref{fig:introfig}).

\section{Pipeline Architecture} \label{sec:pipe}

\begin{figure}[!htb]
    \centering
    \includegraphics[width=0.7\linewidth]{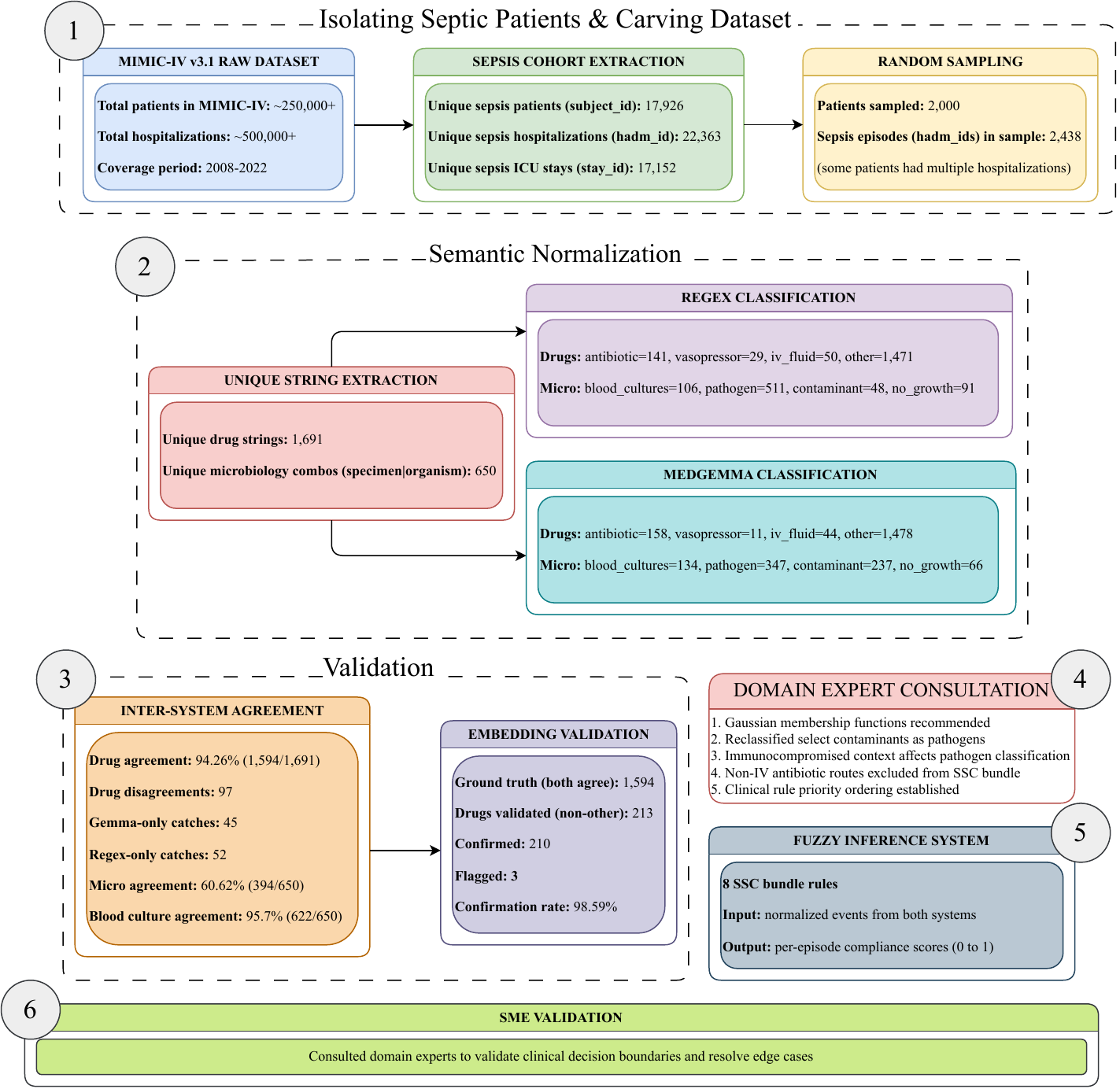}
    \caption{\footnotesize The neuro-symbolic pipeline normalizes clinical text using a dual regex and LLM approach. Following embedding-based validation and domain expert refinement of decision boundaries, a Fuzzy Inference System generates graded compliance scores.}
    \label{fig:pipeline}
\end{figure}

\noindent The Expert-Guided Neuro-Symbolic Pipeline (Figure \ref{fig:pipeline}) integrates semantic normalization, expert validation, and fuzzy reasoning into one workflow, following the hybrid approach in the \hyperref[app:prob_stat]{Appendix: Problem Statement}. We instantiate it on sepsis using $MIMIC-IV$, but the architecture is domain-general and transfers to any protocol expressible as fuzzy rules, given a domain expert to set the decision boundaries. The $MIMIC-IV$ landscape is described in \hyperref[app:data_overview]{Appendix: Data Overview} and the eight SSC rules in \hyperref[app:rules]{Appendix: Eight Rules From SSC}.

\subsection{Cohort Selection and Data Preparation}
\label{subsec:cohort_prep}
\noindent Sepsis episodes were identified from $MIMIC-IV$ using $ICD-9$ codes $038.*$, $995.91$, $995.92$, $785.52$ and $ICD-10$ codes $A40$, $A41$, $R65.20$, $R65.21$, yielding $17,926$ patients across $22,363$ hospitalizations and $17,152$ ICU stays (Figure \ref{fig:pipeline} (1)). Random sampling with seed $55$ produced $2,000$ patients and $2,438$ sepsis episodes ($|S| = 2{,}438$), some patients contributing multiple hospitalizations. The cohort exceeds the power-analysis minimums for Cohen's Kappa ($n = 22$, by $76.9\times$), McNemar's test ($n = 471$, by $5.2\times$), and fuzzy membership estimation ($n = 240$, by $10.2\times$). Each episode is restructured into seven standardized files (medications, microbiology, labs, vitals, static profile), with timestamps preserved to the minute for precise alignment of interventions with protocol windows.

\subsection{Semantic Normalization}
\label{subsec:semantic_norm}
\noindent Clinical text encodes the same drug as a trade name (\texttt{Zosyn}, \texttt{Rocephin}), generic name, abbreviation, or tall-man spelling, and the same specimen as ``blood culture'', ``BC'', or ``blood cx''. The pipeline extracts unique strings once and classifies them centrally, giving $1,691$ unique drug strings and $650$ microbiology combinations, then maps labels back across all episodes. A regex system with domain-informed exclusion rules (filtering topical formulations, flushes, and osmotic saline) produces $141$ antibiotics, $29$ vasopressors, and $50$ IV fluids, while MedGemma-4b-it (4-bit quantized, local execution) under zero-shot structured prompting produces $158$ antibiotics, $11$ vasopressors, and $44$ IV fluids (Figure \ref{fig:pipeline} (2)). Because the LLM is prompted zero-shot and cannot be trusted on its own for clinical labeling, the regex system, which matches against documented drug information, verifies every classification rather than the LLM acting autonomously. The two are kept together because their errors are complementary: regex gives near-perfect precision on explicit matches but zero recall on synonyms, while MedGemma generalizes at the cost of hallucination risk.

\subsection{Validation}
\label{subsec:validation}
\noindent The classifiers agree on $1,594$ of $1,691$ drug strings (94.26\%) and $394$ of $650$ microbiology combinations (60.62\%), with blood-culture detection at 95.7\% (622/650). The $97$ disagreements split into 45 Gemma-only and 52 regex-only catches. Cohen's Kappa gives $\kappa = \frac{P_o - P_e}{1 - P_e} = 0.65$ ($P_o = 0.9426$, $P_e = 0.8277$), substantial on the Landis-Koch scale, and McNemar's test gives $\chi^2 = \frac{(45 - 52)^2}{45 + 52} = 0.505$, $p = 0.088$, failing to reject marginal homogeneity, confirming the classifiers are complementary rather than redundant (confusion matrix in \hyperref[fig:vali_confmat]{Appendix: Drug classification Confusion Matrix}). To rule out shared error, embedding-based validation uses MedGemma's embedding space as an independent reference (Figure \ref{fig:pipeline} (3)): the 1,594 agreed classifications set per-category thresholds ($\theta_{\text{antibiotic}} = 0.3206$, $\theta_{\text{vasopressor}} = 0.2263$, $\theta_{\text{iv\_fluid}} = 0.3075$), against which all 213 non-\texttt{other} Gemma classifications are tested, confirming 210 and flagging 3 ALL-CAPS tall-man tokenizer artifacts ($\text{Confirmation Rate} = 210/213 = 98.59\%$). The 97 disagreements are adjudicated by embedding comparison against both claimed anchors, giving 49 regex wins and 48 Gemma wins (This answers \hyperref[rq2]{RQ2}; see \hyperref[app:rq]{Appendix: Research Question Summary}). Only cases unresolved after the embedding check were escalated as edge cases for clinical review. In total the regex system verifies $220$ clinical drug strings ($141$ antibiotics, $29$ vasopressors, $50$ IV fluids) that a purely synonym-matching system would have missed on variant spellings, with MedGemma resolving the remaining lexical variation. The validated classifications, adjudicated disagreements, and expert-reviewed edge cases (Figure \ref{fig:pipeline}(4)) feed the fuzzy inference system.

\subsection{Fuzzy Compliance Assessment}
\label{subsec:fuzzy_assessment}
\begin{figure}[h]
\centering
\includegraphics[width=\textwidth]{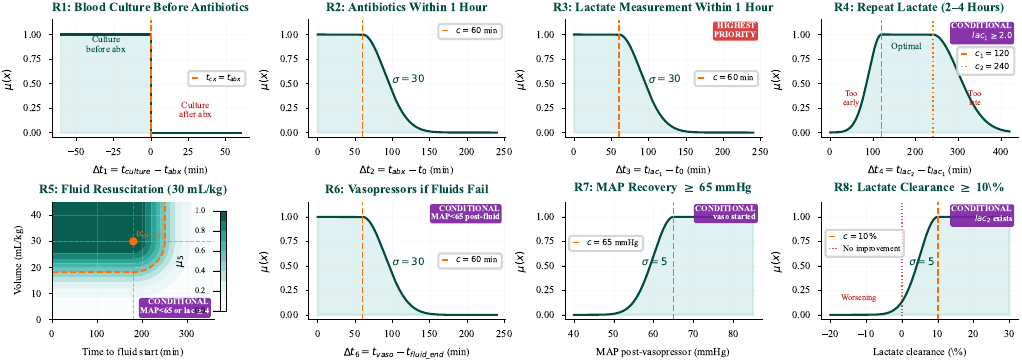}
\caption{\footnotesize Membership functions for all eight SSC bundle rules. Timing-based interventions (Rules 1-3, 6) use Boolean or right-side half-Gaussian decay, while value-based targets (Rules 7-8) utilize left-side half-Gaussians. Rule 4 applies a window penalty, Rule 5 combines volume and time in a 2D contour, and purple badges mark conditional triggers.}
\label{fig:demorulesgraph}
\end{figure}

\noindent Binary scoring misrepresents clinical reality, since a patient treated at $62$ minutes is not categorically different from one treated at $58$. Following \hyperref[app:expert_consult]{Domain Expert Consultation}, the pipeline uses a Sugeno fuzzy inference system (Figure \ref{fig:pipeline} (5)) over eight SSC rules in three phases (Figure \ref{fig:demorulesgraph}). Half-Gaussian membership functions were chosen over triangular or trapezoidal shapes, on expert recommendation, because clinical benefit degrades gradually rather than collapsing at a boundary. Timing-based rules use a right-side half-Gaussian holding $\mu = 1$ inside the window and decaying beyond it; value-based rules use a left-side variant decaying below target; Rule 4 uses a window variant penalizing both early and late repeat lactate.

\begin{equation*}
\mu_{\text{right}}(x;\,c,\sigma) =
\begin{cases}
1 & \text{if } x \leq c \\
\exp\!\left(-\dfrac{(x-c)^{2}}{2\sigma^{2}}\right) & \text{if } x > c
\end{cases}
\end{equation*}

\noindent Phase 1 covers the Hour-1 bundle: Rule 1 is Boolean (cultures before antibiotics), and Rules 2 and 3 apply $\mu_{\text{right}}(\Delta t;\,60,\,30)$, so treatment at 90 minutes yields $\mu = 0.61$ and at 120 minutes $\mu = 0.13$, with Rule 3 (initial lactate) the highest priority and Rule 4 conditional on lactate $\geq 2.0$ mmol/L. Phase 2 covers fluid volume (Rule 5) and vasopressor timing (Rule 6), both conditional on hypotension or severe hyperlactatemia. Phase 3 covers MAP recovery (Rule 7) and lactate clearance (Rule 8), conditional on prior intervention. Missing data is handled by excluding unevaluable conditional rules from the evaluable set $\mathcal{R}_i$, while mandatory Rules 1 to 3 receive $\mu = 0$ when absent. The per-episode score is a weighted Sugeno defuzzification $\text{Compliance}(S_i) = \frac{\sum_{r \in \mathcal{R}_i} R_r \cdot z_r}{\sum_{r \in \mathcal{R}_i} R_r}$, with the expert-set priority $R_3 > R_2 > R_5 > R_6 > R_1 = R_4 = R_7 = R_8$ (see \hyperref[app:fuzval]{Appendix: Fuzzy Membership Function Parameter}). Scores are aggregated across the cohort to surface population-level patterns. All experiments ran locally on a single NVIDIA RTX 5090.

\section{Quantitative Findings with Expert Discussion} \label{sec:results}

\subsection{Regex vs MedGemma Validation}
\label{subsec:regvsgem}

\begin{wrapfigure}{r}{0.45\textwidth}
    \centering
    \vspace{-6pt}
    \includegraphics[width=0.32\textwidth]{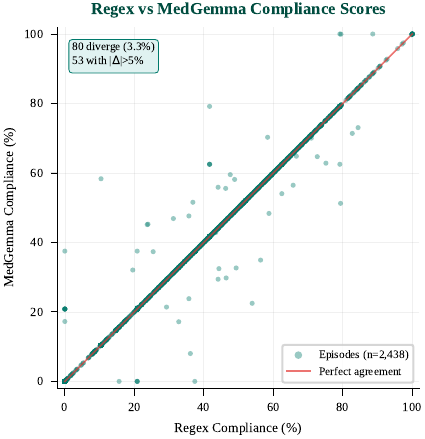}
    \vspace{-10pt}
    \caption{\footnotesize A scatter plot comparing Regex and MedGemma compliance scores, highlighting episodes with divergent classifications.}
    \vspace{-6pt}
    \label{fig:regvsgem}
\end{wrapfigure}
Figure \ref{fig:regvsgem} compares per-episode compliance scores produced by the regex and MedGemma classifiers when used separately across all 2,438 episodes. Since both classifiers agree on 94.26\% of drug classifications, and the fuzzy inference system is fully deterministic given its inputs, episodes where classifiers agree must produce identical compliance scores. This is confirmed in Figure \ref{fig:regvsgem}, where the vast majority of the 2,438 episodes lie exactly on the perfect agreement diagonal. Only 80 episodes (3.3\%) diverge at all, and only 53 of those exhibit score differences exceeding 5\%, corresponding precisely to the adjudicated disagreements where the two classifiers assigned different clinical categories to the same drug string. The regex system verified 220 clinical drug strings against documented drug information, while MedGemma resolved the remaining lexical variation that a purely synonym-matching system would have missed, and the fuzzy inference system then transformed these normalized inputs into graded compliance scores that a purely neural end-to-end model could not guarantee to align with SSC safety boundaries (This answers \hyperref[rq1]{RQ1}; see \hyperref[app:rq]{Appendix: Research Question Summary}).


\subsection{Rule Compliance}
\label{subsec:rulecomp}
\begin{figure}[!htb]
    \centering
    \includegraphics[width=0.60\linewidth]{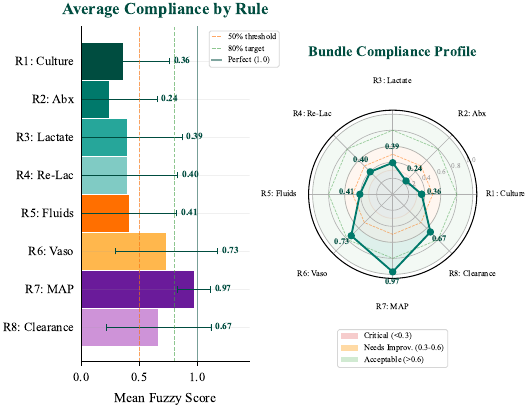}
    \caption{\footnotesize Bar and radar charts displaying average fuzzy compliance scores across all eight bundle rules.}
    \label{fig:rulecomp}
\end{figure}

Figure \ref{fig:rulecomp} reveals a stark stratification across the eight bundle rules. Antibiotic timing (\hyperref[r2]{R2}, $\mu = 0.24$) is the lowest-scoring rule under our operationalization, with only 13\% of episodes meeting the uniform one-hour threshold used in this analysis (Appendix: Figure \ref{fig:08_antibiotic_timing}). This uniform threshold is a cohort-level operational simplification and does not stratify antibiotic timing by septic-shock status; \hyperref[r2]{R2} should therefore be interpreted specifically under this predefined one-hour operationalization. Domain expert discussion (Figure \ref{fig:pipeline} (6)) offered an important clinical explanation: by the time a septic patient arrives in the ICU, clinicians typically already suspect sepsis, and antibiotics are frequently administered prior to formal ICU admission. This is corroborated by Appendix: Figure \ref{fig:08_antibiotic_timing}, where the observed timing distribution shows a substantial proportion of episodes receiving antibiotics before the sepsis onset timestamp recorded in MIMIC-IV, suggesting pre-ICU administration that the pipeline's onset-anchored window penalizes. Blood culture sequencing (\hyperref[r1]{R1}, $\mu = 0.36$) and lactate measurement (\hyperref[r3]{R3}, $\mu = 0.39$) also fall below the 50\% threshold, indicating systemic early-intervention failures. Conditional hemodynamic rules tell a different story: vasopressor initiation (\hyperref[r6]{R6}, $\mu = 0.73$) and MAP recovery (\hyperref[r7]{R7}, $\mu = 0.97$) appear near-acceptable. However, this perceived high performance likely constitutes a \textbf{survivorship bias artifact}, reflecting that once shock is recognized and fluids have failed in a patient who has survived long enough to reach this secondary phase of care, escalation to vasopressors is well executed, and should not be read as genuine compliance excellence. The overall compliance distribution (Figure \ref{fig:01_compliance_distribution}) confirms this systemic underperformance, with a mean of 36.7\% and median of 37.5\% across all episodes, driven primarily by failures in the time-critical Hour-1 bundle components (This answers \hyperref[rq3]{RQ3}; see \hyperref[app:rq]{Appendix: Research Question Summary}).

\subsection{Clinical Cascade}
\label{subsec:cascade}

\begin{wrapfigure}{r}{0.54\textwidth}
    \centering
    \vspace{-6pt}
    \includegraphics[width=0.52\textwidth]{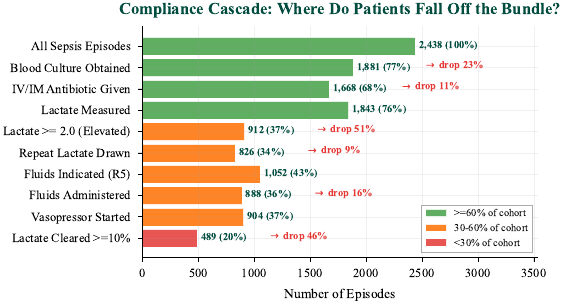}
    \vspace{-10pt}
    \caption{\footnotesize A compliance cascade illustrating patient drop-offs across the sequential sepsis management bundle steps. 
    }
    \vspace{-6pt}
    \label{fig:cascade}
\end{wrapfigure}

Figure \ref{fig:cascade} (the data points here could fall under the first-hour bundle or later in response to follow-up treatment) traces sequential patient drop-offs across the SSC bundle. The most alarming finding, flagged by domain experts, is the 51\% drop at elevated lactate ($\geq 2.0$ mmol/L): only 912 (37\%) episodes recorded this threshold, indicating either unmeasured or undocumented lactate both constituting bundle failures. Appendix: Figure \ref{fig:06_missing_data} corroborates this, showing antibiotic timing (\hyperref[r2]{R2}, 31.6\%) and lactate measurement (\hyperref[r3]{R3}, 24.4\%) as the most data-sparse Hour-1 interventions. Whether these reflect care never delivered or care never recorded, the pipeline conservatively treats both as non-compliance, consistent with SSC guidelines. Downstream conditional rules reflect this sparsity, with fluid resuscitation (\hyperref[r5]{R5}) activating in only 43\% of episodes and lactate clearance in 34\% (Appendix: Figure \ref{fig:05_activation_rates}).


\subsection{ICU Outcomes}
\label{subsec:icu}

\begin{figure}[!htb]
    \centering
    \includegraphics[width=0.50\linewidth]{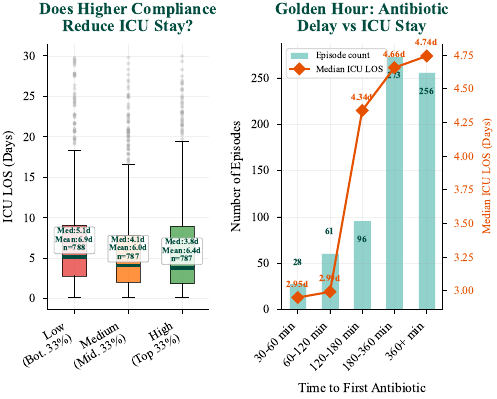}
    \caption{\footnotesize Descriptive comparison of ICU length of stay across overall bundle-compliance groups and antibiotic-timing categories.}
    \label{fig:icu}
\end{figure}

Figure \ref{fig:icu} reinforces the critical importance of early intervention, showing that episodes where antibiotics are administered within 30–60 minutes of onset achieve a median ICU stay of just 2.95 days. This duration rises sharply to 4.74 days when administration is delayed beyond six hours. This primarily demonstrates that low-compliance episodes result in a median stay of 5.1 days, whereas the high-compliance group averages only 3.8 days. The broader clinical implications of these findings are detailed in the appendix. Appendix: Figure \ref{fig:09_los_vs_compliance} confirms the consistent association between higher bundle compliance and shorter ICU stays ($r = -0.026$, $n = 2,362$), with a trend showing a reduction of 0.006 days per 1\% gain in compliance. Furthermore, Appendix: Figure \ref{fig:10_los_vs_worst_rule} localizes the clinical burden to specific metrics: episodes where the worst-performing rule is antibiotic timing (\hyperref[r2]{R2}, median 7.0 days) or lactate measurement (\hyperref[r3]{R3}, median 5.0 days) carry the highest ICU costs. Collectively, these data establish early Hour-1 intervention failures as the primary driver of prolonged critical care (This answers \hyperref[rq4]{RQ4}; see \hyperref[app:rq]{Appendix: Research Question Summary}).

\subsection{Operational Insights}
\label{subsec:opins}
Median compliance rises from $0.00$ in the Hour-1 bundle to $0.67$ in resuscitation and $1.00$ in treatment response, confirming that immediate interventions constitute the dominant failure mode across all three clinical phases (Appendix: Figure \ref{fig:phases}). Of the $2{,}000$ patients, $14\%$ experienced recurrent sepsis, yet median compliance improves only marginally from $0.362$ on the first episode to $0.375$ on the second ($+1.3\%$), suggesting that neither clinical learning nor institutional feedback loops produce meaningful compliance gains across repeated admissions for the same patient (Appendix: Figure \ref{fig:patientvsepi}). 
\vspace{-2mm}




\section{Conclusion} \label{sec:conclusion}


In this paper, we presented an Expert-Guided Neuro-Symbolic Pipeline that demonstrates combining semantic normalization, fuzzy reasoning, and domain expertise produces interpretable and graded compliance assessments for sepsis care. Our \hyperref[app:combres]{Results} and \hyperref[app:companal]{Comparison} revealed systemic Hour-1 compliance failures alongside descriptive differences in ICU length of stay across compliance and antibiotic-timing groups. This pipeline is not limited to sepsis and can be adapted to other clinical protocols such as stroke management or cardiac care, provided the underlying guidelines (fuzzy rules) are well-defined and a domain expert is available to set the boundaries. Several limitations point to future work. Mortality was outside the scope of this study, which deliberately measured how closely the sepsis bundle was followed rather than its survival impact; the high conditional-rule scores we report reflect a survivorship-bias artifact rather than outcome modeling, and linking graded compliance to mortality is a natural next step. The onset-anchored timing window penalizes antibiotics delivered before ICU admission, so future versions should acknowledge that sepsis care often begins earlier and credit early pre-ICU antibiotics accordingly. The fuzzy boundaries are currently fixed from expert input alone; a future data-plus-expert calibration step, optimizing the membership parameters against outcome data while preserving the expert-set rule priorities, would improve robustness and reduce manual tuning. Finally, the pipeline depends on subject matter experts to establish clinically meaningful decision boundaries, which may limit scalability in resource-constrained settings. As generative models continue to mature in medical reasoning, they may progressively reduce this dependency, enabling more autonomous and widely deployable clinical decision support across diverse healthcare domains.

\bibliography{ref}

\appendix

\section{Data Availability and Ethical Statement}
Data Availability and Ethical Statement
This study uses MIMIC-IV v3.1, a de-identified critical care database sourced from the electronic health records of Beth Israel Deaconess Medical Center (BIDMC), spanning 2008–2022 and covering approximately 250,000 patients across 500,000 hospitalizations. MIMIC-IV is publicly available through PhysioNet \citep{MM4} but constitutes credentialed-access data: access requires completion of a recognized human subjects research training program and execution of a PhysioNet Credentialed Health Data Use Agreement (DUA), which prohibits re-identification of individuals, redistribution of the data, and non-research use.

\noindent The Institutional Review Board (IRB) at BIDMC granted a waiver of informed consent and approved the sharing of MIMIC-IV as a research resource; no additional IRB approval was required for this study. All patient identifiers were removed in accordance with the HIPAA Safe Harbor de-identification standard prior to public release. The raw MIMIC-IV data, derived cohorts, and extracted episode files used in this pipeline cannot be shared directly by the authors and must be independently obtained through PhysioNet.

\section{Eight Rules From SSC}\label{app:rules}
\noindent The following are the eight rules from the SSC \citep{SSC} that govern the compliance assessment in this pipeline.

\begin{enumerate}
    \item \textbf{[$R1$] Blood Cultures Before Antibiotics: } \label{r1} Preserve pathogen identification before sterilizing the blood.
    \begin{quote}
    \textit{``Obtain blood cultures before administering antibiotics.''}
    \end{quote}

    \item \textbf{[$R2$] Broad-Spectrum Antibiotics: } \label{r2} Initiate empirical antibiotic coverage without delay.
    \begin{quote}
    \textit{``Administer broad-spectrum antibiotics.''}
    \end{quote}

    \item \textbf{[$R3$] Measure Lactate: } \label{r3} Assess tissue hypoperfusion severity immediately upon sepsis onset.
    \begin{quote}
    \textit{``Measure lactate level.''}
    \end{quote}

    \item \textbf{[$R4$] Re-measure Lactate if High: } \label{r4} Confirm persistent hypoperfusion when initial lactate is elevated.
    \begin{quote}
    \textit{``Remeasure lactate if initial lactate is elevated ($>$ 2 mmol/L).''}
    \end{quote}

    \item \textbf{[$R5$] IV Fluid Resuscitation: } \label{r5} Restore circulating volume in hemodynamically compromised patients.
    \begin{quote}
    \textit{``Begin rapid administration of 30 mL/kg crystalloid for hypotension or lactate $\geq$ 4 mmol/L.''}
    \end{quote}

    \item \textbf{[$R6$] Vasopressors: } \label{r6} Support perfusion pressure when fluids alone are insufficient.
    \begin{quote}
    \textit{``Apply vasopressors if hypotensive during or after fluid resuscitation to maintain a mean arterial pressure (MAP) $\geq$ 65 mm Hg.''}
    \end{quote}

    \item \textbf{[$R7$] MAP Target: } \label{r7} Target minimum perfusion pressure to prevent organ damage.
    \begin{quote}
    \textit{``For adults with septic shock on vasopressors, we recommend an initial target mean arterial pressure (MAP) of 65 mm Hg over higher MAP targets.''}
    \end{quote}

    \item \textbf{[$R8$] Lactate Clearance: } \label{r8} Verify adequate metabolic response to resuscitation efforts.
    \begin{quote}
    \textit{``For adults with sepsis or septic shock, we suggest guiding resuscitation to decrease serum lactate in patients with elevated lactate levels over not using serum lactate.''}
    \end{quote}
\end{enumerate}

\noindent \textbf{Note:} The 10\% lactate \hyperref[r8]{Rule 8} clearance threshold does not originate from the SSC bundle directly but is derived from \cite{jones2010lactate}.

\section{Notation and Symbols for Problem Statement}\label{tab:symbols}

\begin{table}[H]
\centering
\caption{\footnotesize Notation and Symbols used to formulate the \hyperref[app:prob_stat]{Problem Statement}}
\begin{tabular}{lp{10cm}}
\toprule
\textbf{Symbol} & \textbf{Definition} \\
\midrule
$v$ & Documented drug variant in clinical records \\
$V$ & Predefined synonym set in rule-based system \\
$\mathcal{L}$ & Large Language Model for semantic normalization \\
$\text{input}_m$ & Clinical prompt supplied to $\mathcal{L}$ \\
$S_k$ & Sepsis episode $k$ \\
$\mathbb{P}$ & $\mathbb{P}(E)$ denotes probability of event $E$ \\
$\mathcal{D}$ & Clinical decision boundary (expert-validated threshold) \\
$\mathcal{F}$ & Fuzzy Inference System encoding SSC logic \\
\bottomrule
\end{tabular}
\end{table}

\section{Problem Statement} \label{app:prob_stat}

\noindent The Surviving Sepsis Campaign guidelines specify critical actions within narrow time windows: antibiotic administration, fluid resuscitation, lactate measurement, and blood cultures. Systematic compliance evaluation across large cohorts remains infeasible because clinical documentation encodes drug names as trade names, generic names, and abbreviations interchangeably, while microbiology results intermix suspected organisms, confirmed pathogens, and contamination events, creating systematic barriers to assessment. Traditional rule-based systems fail under this complexity. When a clinician documents \texttt{Zosyn}, a rule searching for \texttt{piperacillin-tazobactam} fails due to syntactic rigidity, where \hyperref[tab:symbols]{$v$} denotes the documented variant and \hyperref[tab:symbols]{$V$} the predefined synonym set:

\begin{equation*}
\text{Recognize}(v,\, V) =
\begin{cases}
1 & \text{if } v \in V \\
0 & \text{if } v \notin V
\end{cases}
\end{equation*}

\noindent Large Language Models \hyperref[tab:symbols]{$\mathcal{L}$} resolve this variability but introduce unquantified hallucination risk. The probability \hyperref[tab:symbols]{$\mathbb{P}$}, that the neural decisions will result in verifiable grounding in documented facts is undefined:

\begin{equation*}
\text{Determinism}(\mathcal{L}(\text{input}_m)) = \mathbb{P}\!\left(\mathcal{L}(\text{input}_m) \text{ grounded in verifiable facts} \mid \text{input}_m\right) \rightarrow  \text{undefined}
\end{equation*}

\noindent where \hyperref[tab:symbols]{$\text{input}_m$} denotes the clinical prompt supplied to \hyperref[tab:symbols]{$\mathcal{L}$}. No existing pipeline produces graded compliance assessment by combining semantic normalization with transparent symbolic reasoning. We propose an Expert-Guided Neuro-Symbolic Pipeline that constrains \hyperref[tab:symbols]{$\mathcal{L}$} strictly to semantic normalization, preventing autonomous decision-making. A Fuzzy Inference System \hyperref[tab:symbols]{$\mathcal{F}$} then applies expert-validated decision boundaries \hyperref[tab:symbols]{$\mathcal{D}$} to episode \hyperref[tab:symbols]{$S_k$} to produce graded compliance scores:

\begin{equation*}
\text{Compliance}(S_k,\, \mathcal{D}) = \mathcal{F}\!\left(\text{normalized}(S_k),\, \mathcal{D}\right)
\end{equation*}

\noindent These scores generate actionable insights for healthcare professionals across 
$2{,}438$ sepsis episodes from the $MIMIC-IV$ database.

\section{Data Overview}\label{app:data_overview}
\begin{figure}

    \centering
    \includegraphics[width=1\linewidth]{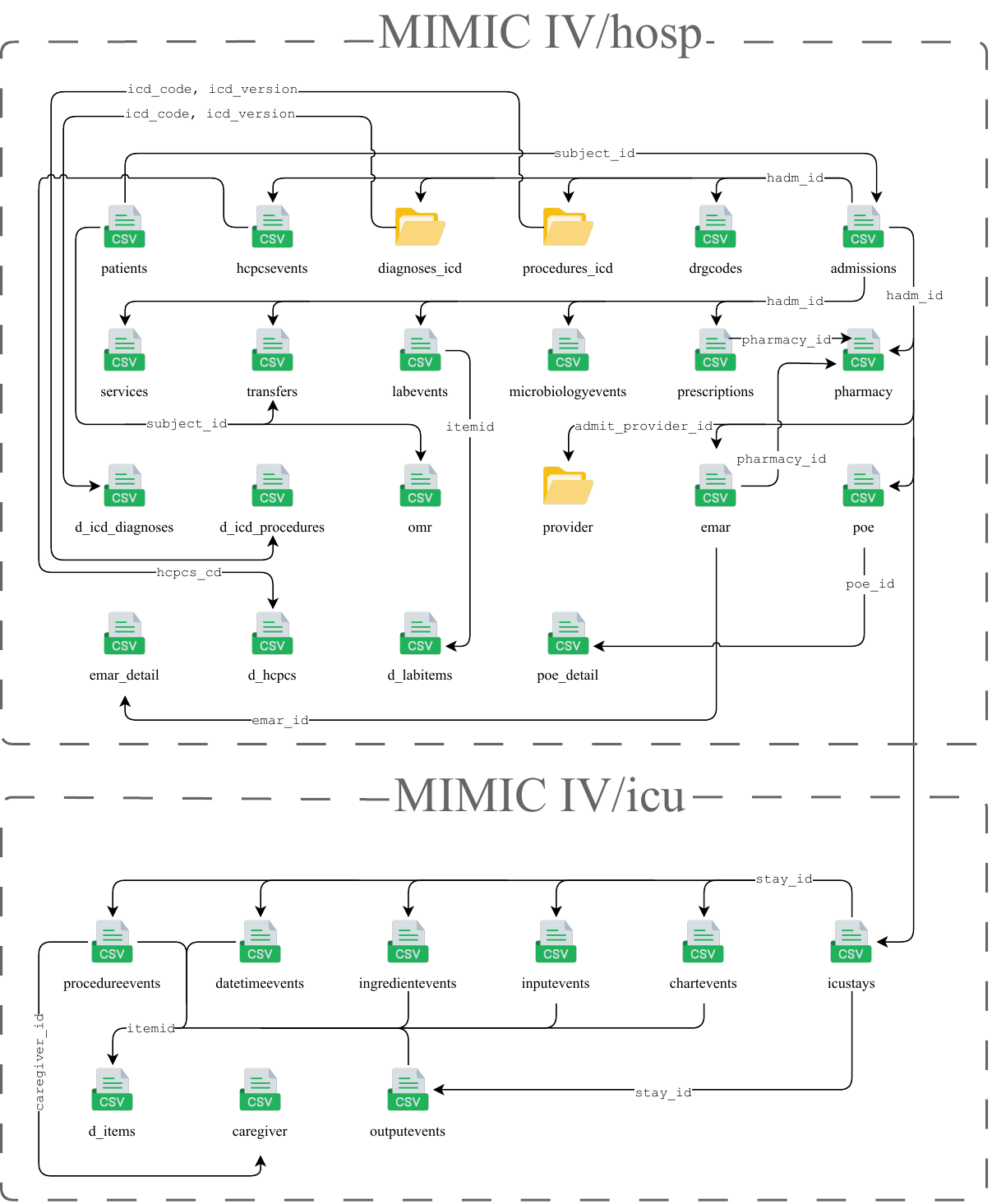}
    \caption{\footnotesize MIMIC-IV Database Structure and Relational Organization. The database comprises two modules: MIMIC-IV/hosp maintaining hospital-level records indexed by \texttt{subject\_id}, \texttt{hadm\_id}, and \texttt{pharmacy\_id}, and MIMIC-IV/icu (9 tables, bottom portion) maintaining ICU-specific records indexed by \texttt{stay\_id}. Primary linking identifiers (\texttt{subject\_id}, \texttt{hadm\_id}, \texttt{stay\_id}) enable integration across administrative, clinical, and intensive care data layers. The pipeline extracts data from \texttt{emar}, \texttt{prescriptions}, \texttt{microbiologyevents}, \texttt{labevents}, \texttt{inputevents}, \texttt{chartevents}, and \texttt{icustays}.}
    \label{fig:dataview}
\end{figure}

The pipeline operates on the $MIMIC-IV v3.1$ \citep{MM4} database, a comprehensive critical care dataset spanning $2008-2022$ containing electronic health records for approximately $250,000$ patients across $500,000+$ hospitalizations. $MIMIC-IV$ is organized into two relational modules. $MIMIC-IV/hosp$, which maintains hospital-level administrative and clinical data, and $MIMIC-IV/icu$, which contains intensive care unit-specific measurements and events. As depicted in Figure \ref{fig:dataview}, these modules share three primary linking identifiers enabling integration across granularities: 1. \texttt{subject\_id} (unique patient), 2. \texttt{hadm\_id} (unique hospitalization), and 3. \texttt{stay\_id} (unique ICU stay). The $MIMIC-IV/hosp$ module comprises $22$ tables organized hierarchically. Patient-level tables include \texttt{patients} and \texttt{admissions}. Clinical tables indexed by \texttt{hadm\_id} include \texttt{diagnoses\_icd} and \texttt{procedures\_icd} for coded diagnoses and procedures, \texttt{labevents} for laboratory measurements, \texttt{microbiologyevents} for culture results, and \texttt{prescriptions} and \texttt{pharmacy} for medication records. The \texttt{emar} (electronic medication administration record) and \texttt{poe} (provider order entry) tables capture medication administration events at granular timestamps. Reference tables (\texttt{d\_icd\_diagnoses}, \texttt{d\_icd\_procedures}, \texttt{d\_hcpcs}, \texttt{d\_labitems}) provide standardized mappings. The $MIMIC-IV/icu$ module contains 9 tables keyed by \texttt{stay\_id}. The \texttt{icustays} table anchors ICU admission records. Measurement tables include \texttt{chartevents} (bedside charted observations), \texttt{inputevents} (medications and fluids administered), \texttt{outputevents} (fluid excretion), \texttt{procedureevents} (procedures performed), and \texttt{datetimeevents} (timestamped clinical events). The \texttt{ingredientevents} table provides medication component granularity. Reference table \texttt{d\_items} maps \texttt{itemid} codes to clinical concepts. For the Surviving Sepsis Campaign compliance assessment, the pipeline specifically extracts data from seven critical tables. From $MIMIC-IV/hosp$, \texttt{prescriptions} and \texttt{emar} capture medication administration timing and names, \texttt{microbiologyevents} provides culture specimen types and interpreted organisms, and \texttt{labevents} contains lactate and other laboratory measurements. From $MIMIC-IV/icu$, \texttt{inputevents} records fluids and vasopressors, \texttt{chartevents} captures vital signs including blood pressure, and \texttt{icustays} anchors the ICU timeline. This selection reflects SSC bundle requirements: 1. Antibiotics (medications), 2. Vasopressors, 3. IV fluids (inputevents), 4. Blood cultures (microbiologyevents), 5. Lactate measurement (labevents), 6. Hemodynamic targets (chartevents vital signs). The relational structure presents both opportunities and challenges. Integration across \texttt{subject\_id}, \texttt{hadm\_id}, and \texttt{stay\_id} enables linking patient demographics with clinical events across hospitalizations. However, unstructured text appears in multiple locations: medication names in \texttt{prescriptions} vary as trade names, generic names, and abbreviations. This heterogeneity necessitates the semantic normalization component described in Section \ref{subsec:semantic_norm}.


\section{Domain Expert Consultation} \label{app:expert_consult}
Subject Matter Experts (SME) (e.g. Clinician) were integrated in the middle of the pipeline design, not retrospectively. Domain clinicians reviewed classifier outputs with the entire pipeline and provided four critical inputs:

\begin{enumerate}
    \item Recommending Gaussian membership functions (reflecting gradual transitions rather than sharp boundaries in clinical adherence)
    \item Reclassifying organisms based on immunocompromised status (e.g., Coagulase-negative Staphylococcus is pathogen in immunocompromised patients but contaminant in immunocompetent)
    \item Mandating route-specific exclusion rules (only IV/IM antibiotics within time windows count toward SSC compliance, excluding topical and oral prophylaxis)
    \item Establishing rule priority ordering (Rule 3: Lactate $>$ Rule 2: Antibiotics $>$ Rule 5: Fluids $>$ Rule 6: Vasopressors $>$ Remaining rules).
\end{enumerate}

\noindent These decisions shaped both normalization rules and fuzzy parameters. With expert-validated parameters established, the normalized events are evaluated through a fuzzy inference system that applies these clinical boundaries to produce graded compliance assessments.

\section{Fuzzy Membership Function Parameter} \label{app:fuzval}
\begin{table}[!h]
\caption{Membership function parameters for the eight SSC bundle compliance rules. Boundaries ($c$) and decay rates ($\sigma$) were established through domain expert consultation. Rules marked $^\dagger$ are conditional and activate only when the clinical trigger is met.}
\label{tab:fuzzy-mf}
\centering
\scriptsize
\setlength{\tabcolsep}{3pt}
\begin{tabular}{@{}lllll@{}}
\toprule
\textbf{Rule} & \textbf{Target} & \textbf{MF Type} & \textbf{Parameters} & $\boldsymbol{\text{Rule Weight}}$ \\
\midrule
\multicolumn{5}{@{}l}{\textit{Phase 1: Hour-1 Bundle}} \\
R1 & Culture before abx & Boolean step & $\mu{=}1$ if $t_{\text{cx}}{<}t_{\text{abx}}$, else 0 & 0.5 \\
R2 & Abx $\leq$1\,hr & Right HG & $c{=}60$\,min, $\sigma{=}30$ & 0.9 \\
R3 & Lactate $\leq$1\,hr & Right HG & $c{=}60$\,min, $\sigma{=}30$ & \textbf{1.0} \\
R4$^\dagger$ & Re-lactate 2--4\,hr & Window HG & $c_1{=}120$, $c_2{=}240$\,min & 0.5 \\
 & lac$_1{\geq}2.0$ & & $\sigma_e{=}30$, $\sigma_l{=}60$ & \\
\midrule
\multicolumn{5}{@{}l}{\textit{Phase 2: Hemodynamic Resuscitation}} \\
R5$^\dagger$ & Fluids 30\,mL/kg & Product & $c_t{=}180$\,min, $\sigma_t{=}60$ & 0.8 \\
 & MAP${<}65$ or lac${\geq}4$ & $\mu_{5t}{\times}\mu_{5v}$ & $c_v{=}30$\,mL/kg, $\sigma_v{=}10$ & \\
R6$^\dagger$ & Vasopressors & Right HG & $c{=}60$\,min, $\sigma{=}30$ & 0.7 \\
 & MAP${<}65$ post-fluid & & & \\
\midrule
\multicolumn{5}{@{}l}{\textit{Phase 3: Treatment Response}} \\
R7$^\dagger$ & MAP $\geq$65\,mmHg & Left HG & $c{=}65$\,mmHg, $\sigma{=}5$ & 0.5 \\
R8$^\dagger$ & Clearance $\geq$10\% & Left HG & $c{=}10\%$, $\sigma{=}5$ & 0.5 \\
\bottomrule
\end{tabular}
\end{table}

\begin{table}[!h]
\caption{Sugeno fuzzy system properties and missing data handling. Priority weights encode the SME-established ordering $Rule_3{>}Rule_2{>}Rule_5{>}Rule_6{>}Rule_1{=}Rule_4{=}Rule_7{=}Rule_8$.}
\label{tab:fuzzy-system}
\centering
\scriptsize
\setlength{\tabcolsep}{3pt}
\begin{tabular}{@{}ll@{}}
\toprule
\textbf{Property} & \textbf{Value} \\
\midrule
Inference type & Sugeno first-order \\
Defuzzification & $\sum_{r \in R_i} R_r z_r \,/\, \sum_{r \in R_i} R_r$ \\
Conditional untriggered & $\mu = 1.0$, excluded from $R_i$ \\
Missing mandatory (R1--R3) & $\mu = 0$ \\
Culture only, no abx (R1) & $\mu = 0.5$ \\
No weight available (R5) & $\mu_{5v} = 0.5$ penalty \\
Missing trigger data & Excluded from $R_i$ \\
Classification & Consensus: agree $\cup$ (embed $\geq \theta$) $\cup$ SME \\
\bottomrule
\end{tabular}
\end{table}

\section{Drug classification Confusion Matrix}\label{fig:vali_confmat}
\begin{figure}[H]
\centering
\includegraphics[width=0.75\columnwidth]{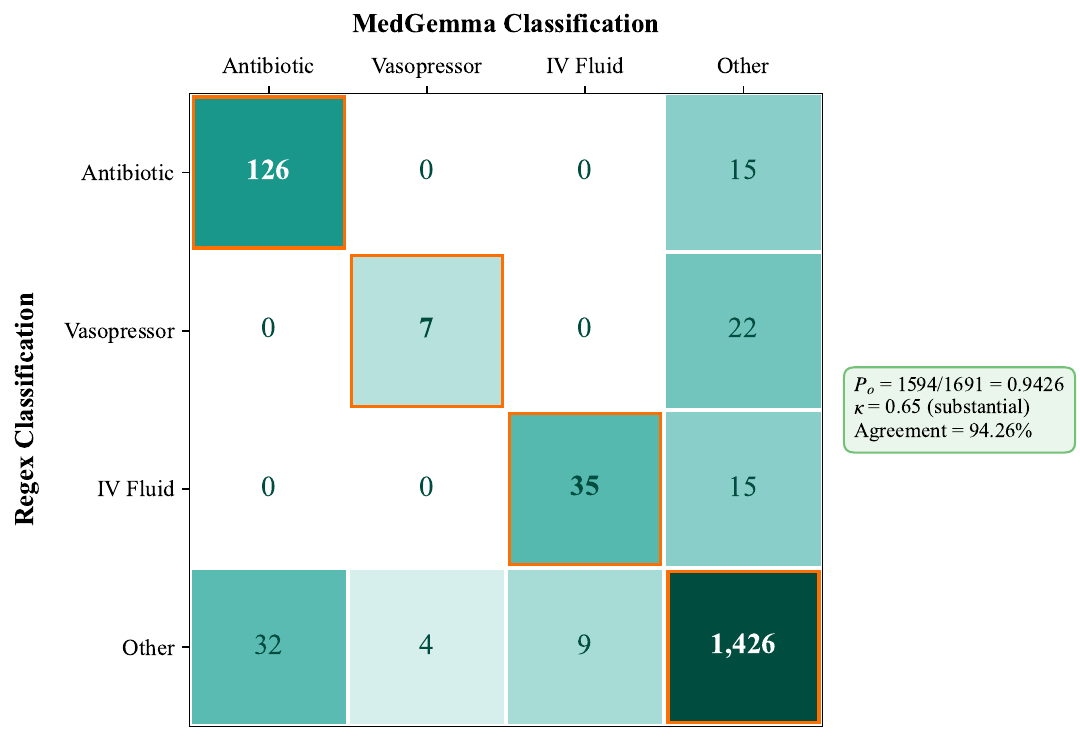}
\caption{\footnotesize Drug classification confusion matrix between regex-based and MedGemma classifiers 
across 1,691 unique drug strings. Diagonal cells (orange borders) represent agreement ($P_o = 1{,}594/1{,}691 = 0.9426$). Off-diagonal entries reveal complementary error profiles: regex misses 32 antibiotics, 4 vasopressors, and 9 IV fluids that MedGemma captures, while MedGemma misses 15 antibiotics, 22 vasopressors, and 15 IV fluids that regex captures. Cohen's $\kappa = 0.65$ (substantial agreement). All off-diagonal cells within the clinical-category submatrix (antibiotic, vasopressor, IV fluid) are zero, confirming that disagreements occur exclusively between a clinical category and \texttt{other}, never across clinical categories.}
\end{figure}

\clearpage
\section{Additional Graphs}\label{app:addgraph}
\begin{figure}[H]
    \centering
    \includegraphics[width=0.75\linewidth]{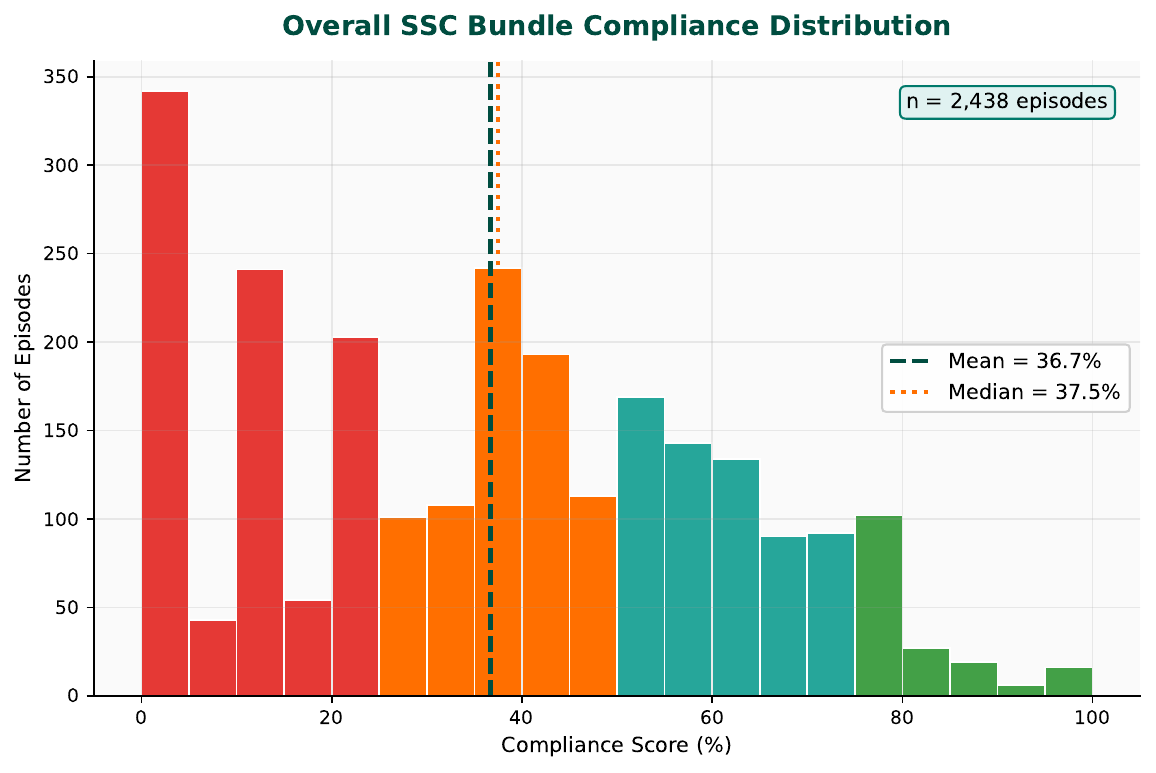}
    \caption{\footnotesize Overall distribution of fuzzy compliance scores across the cohort of 2,438 sepsis episodes. The histogram is color-coded by compliance quartile, revealing a heavy concentration of episodes in the lower performance tiers. The mean overall compliance score is 36.7\% (median 37.5\%), highlighting significant systemic challenges in executing the complete set of Surviving Sepsis Campaign (SSC) bundle guidelines in clinical practice.}
    \label{fig:01_compliance_distribution}
\end{figure}

\begin{figure}[H]
    \centering
    \includegraphics[width=0.75\linewidth]{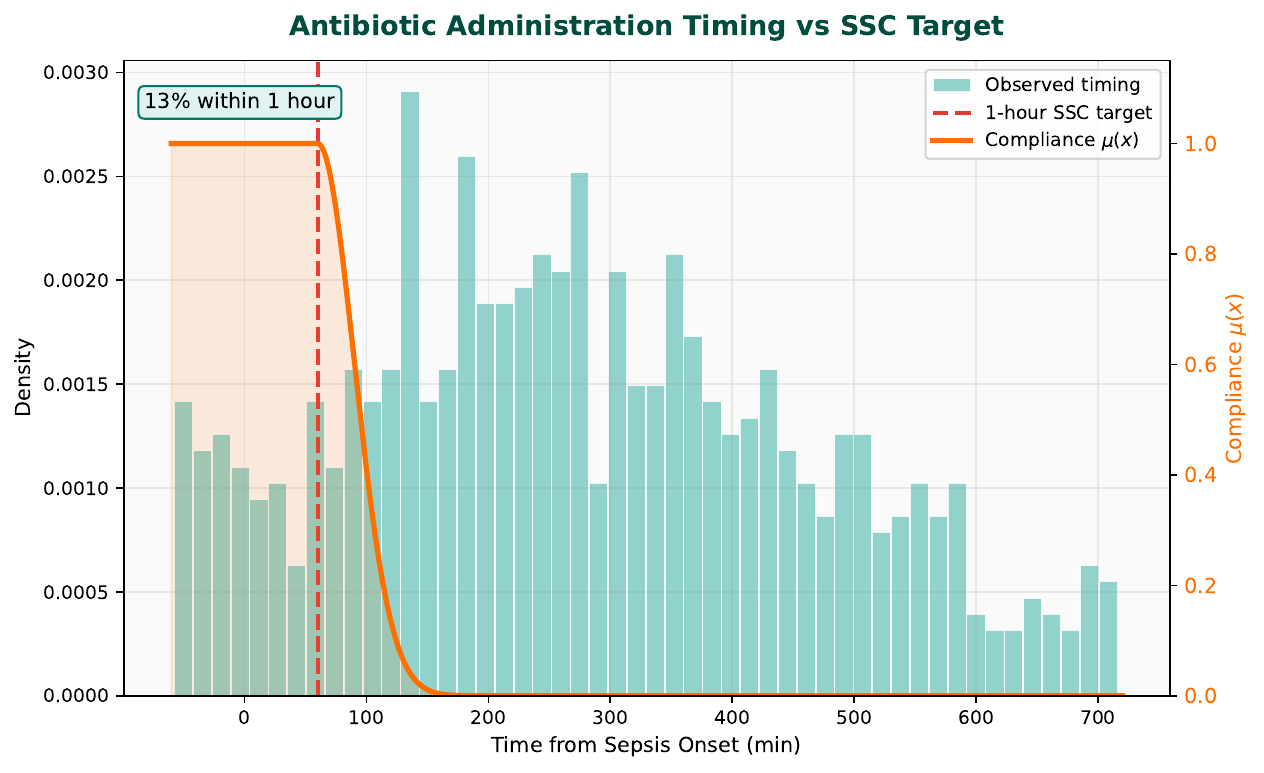}
    \caption{\footnotesize Distribution of time from sepsis onset to initial antibiotic administration, overlaid with the fuzzy compliance membership function $\mu(x)$ (orange line). A severe operational bottleneck is evident, as only 13\% of the observed episodes met the strict 1-hour SSC target (red dashed line).}
    \label{fig:08_antibiotic_timing}
\end{figure}

\begin{figure}[H]
    \centering
    \includegraphics[width=0.75\linewidth]{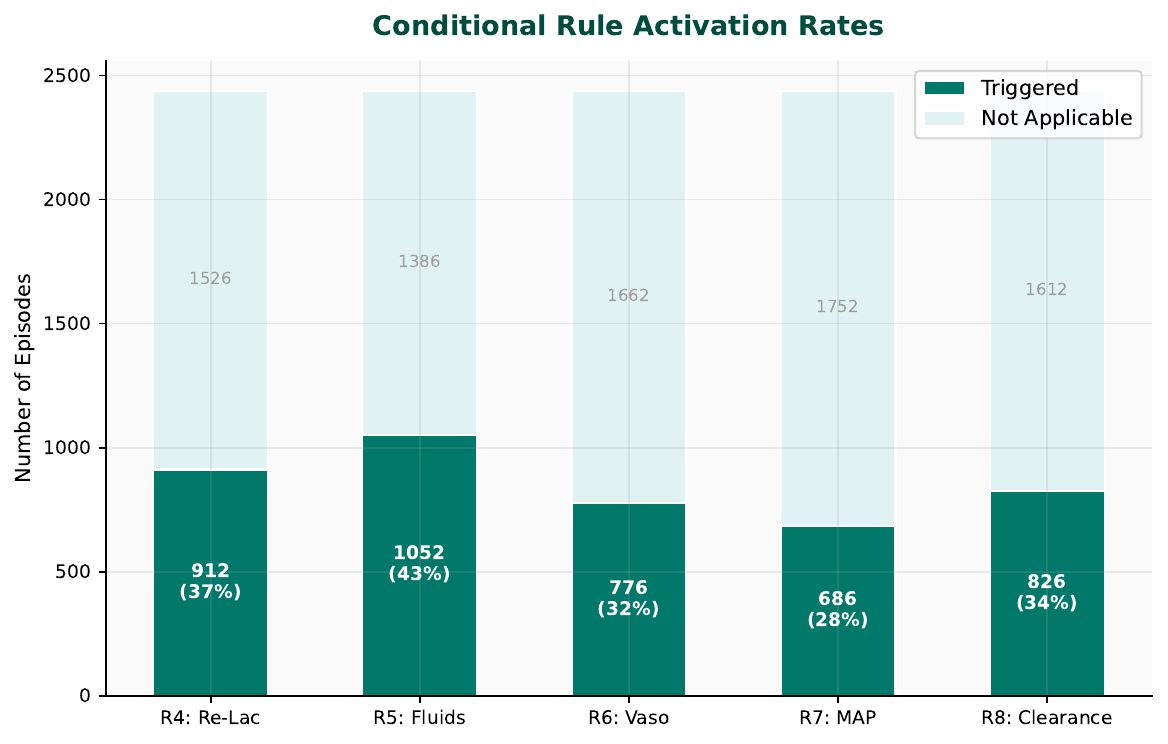}
    \caption{\footnotesize Proportion of episodes triggering the conditional rules (R4 through R8) of the SSC bundle. This stacked chart visualizes clinical acuity; fluid resuscitation (R5) was the most frequently activated conditional intervention (43\% of episodes), reflecting a high prevalence of shock or severe hyperlactatemia in the cohort.}
    \label{fig:05_activation_rates}
\end{figure}

\begin{figure}[H]
    \centering
    \includegraphics[width=0.75\linewidth]{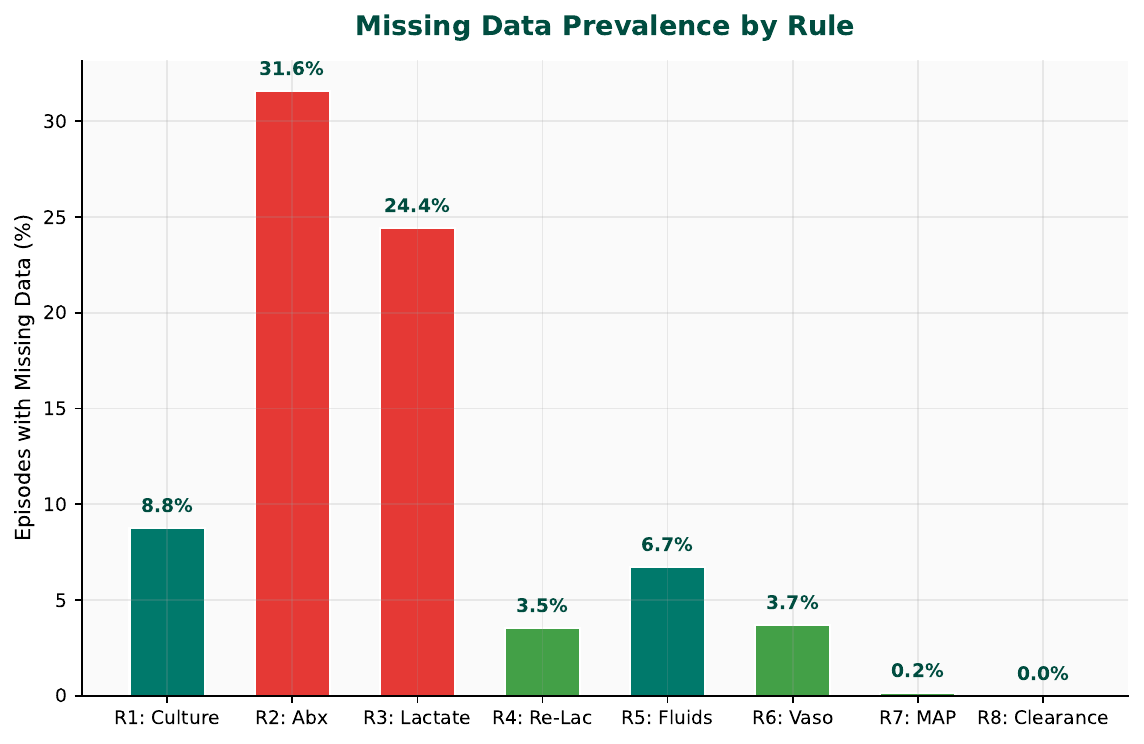}
    \caption{\footnotesize Prevalence of missing clinical documentation across the eight SSC bundle components. Data sparsity is most severe during the critical early stages of care, specifically regarding the exact timing of antibiotic administration (R2, 31.6\%) and initial lactate measurements (R3, 24.4\%).}
    \label{fig:06_missing_data}
\end{figure}

\begin{figure}
    \centering
    \includegraphics[width=0.65\linewidth]{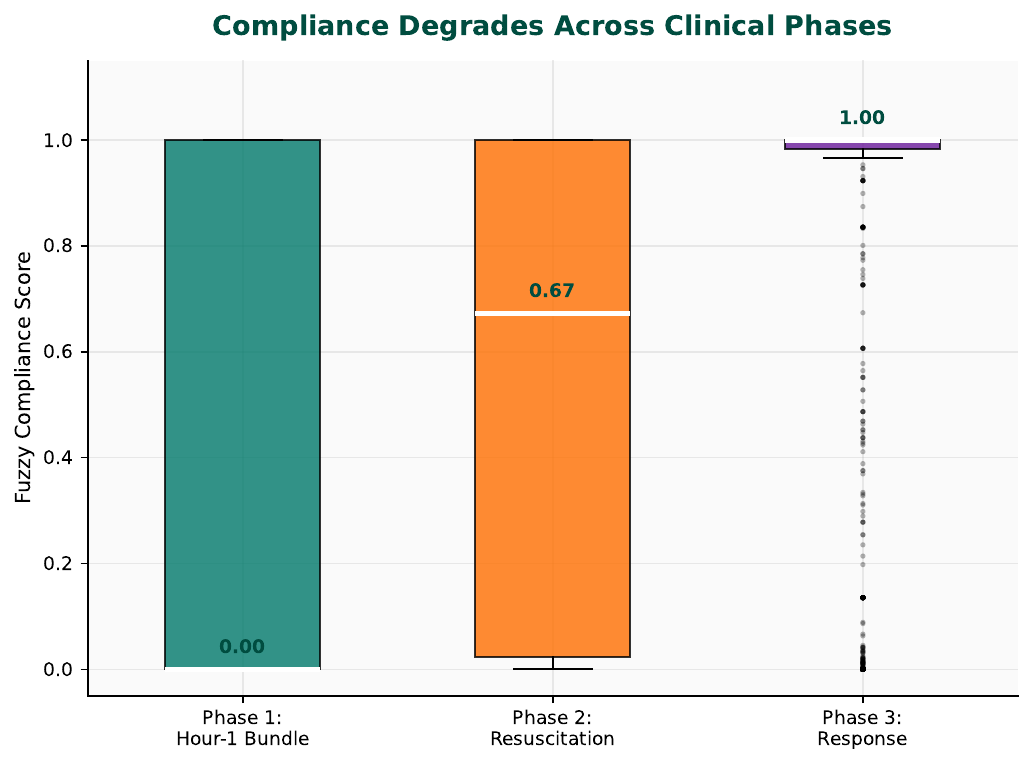}
    \caption{\footnotesize Box plots illustrating the distribution of fuzzy compliance scores categorized by clinical care phase: Phase 1 (Hour-1 Bundle), Phase 2 (Resuscitation), and Phase 3 (Response). The data demonstrates a significant increase in median compliance from Phase 1 (0.00) to Phase 3 (1.00). This high variance highlights the operational difficulty of executing rapid, immediate interventions in the initial acute phase versus achieving compliance in downstream resuscitation and response metrics.}
    \label{fig:phases}
\end{figure}

\begin{figure}
    \centering
    \includegraphics[width=0.65\linewidth]{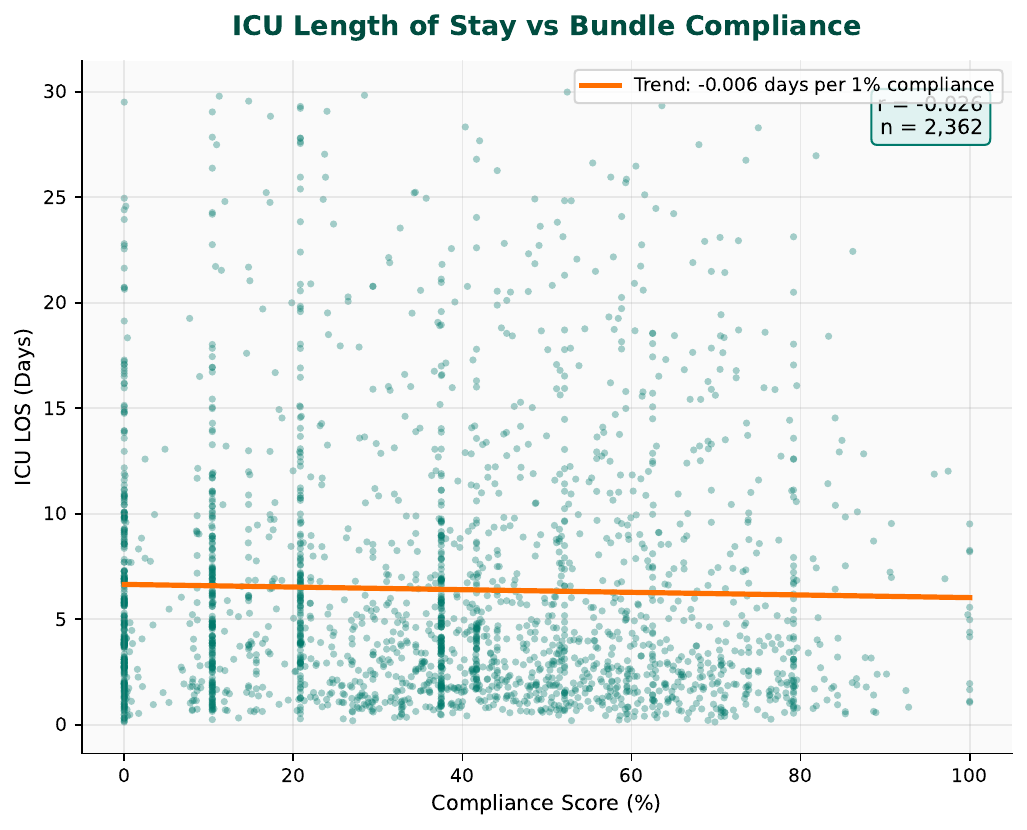}
    \caption{\footnotesize This scatter plot evaluates the correlation between overall fuzzy compliance scores and Intensive Care Unit Length of Stay (ICU Length of Stay (LOS)). The orange trend line demonstrates that \textbf{higher bundle compliance leads to a reduction in ICU time}. Specifically, the analysis shows a trend of \textbf{-0.006 days} of ICU stay saved for every \textbf{1\% increase in compliance}, based on a correlation of $r = -0.026$ across $n = 2,362$ episodes.}
    \label{fig:09_los_vs_compliance}
    \label{fig:loscomp}
\end{figure}

\begin{figure}[H]
    \centering
    \includegraphics[width=0.75\linewidth]{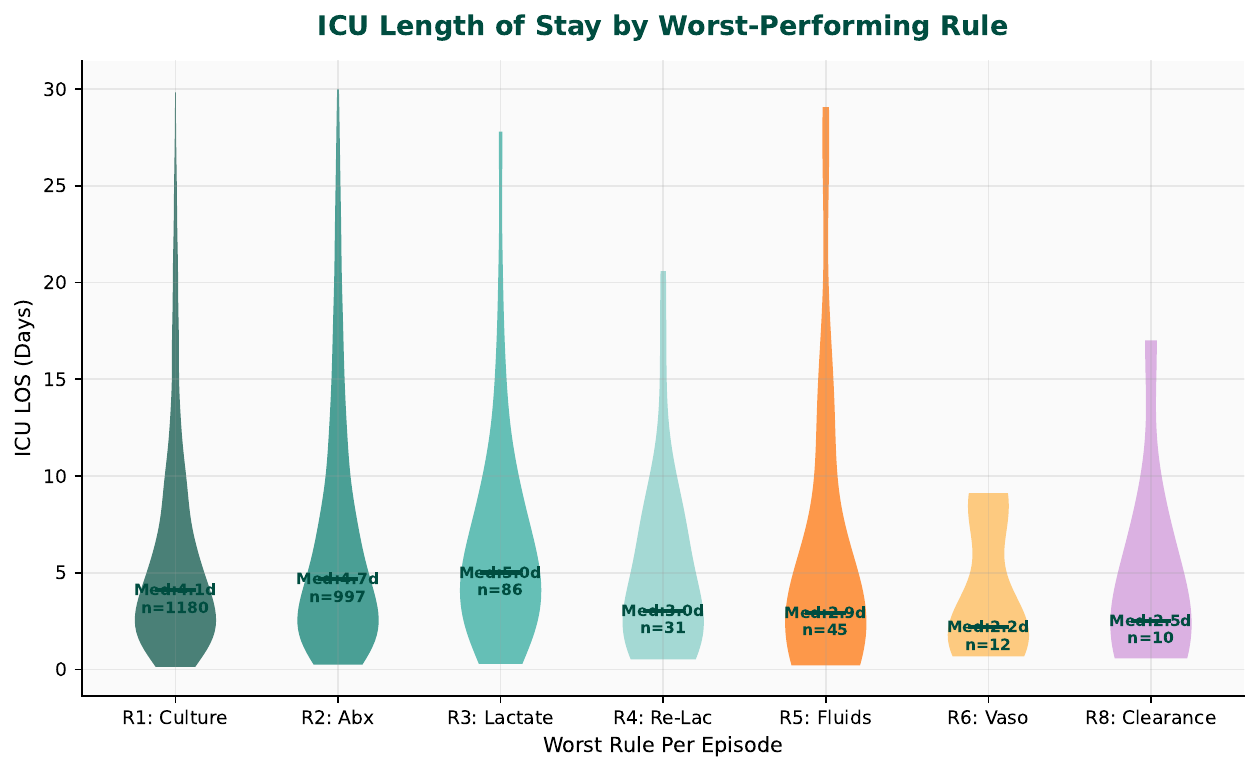}
    \caption{\footnotesize Violin plots displaying ICU Length of Stay distributions, grouped by the specific bundle rule that yielded the lowest compliance score per episode. Failures in early interventions such as delayed antibiotics (R2, median 7.0 days) and delayed initial lactate (R3, median 5.0 days) are strongly associated with prolonged ICU stays.}
    \label{fig:10_los_vs_worst_rule}
\end{figure}

\begin{figure}
    \centering
    \includegraphics[width=0.75\linewidth]{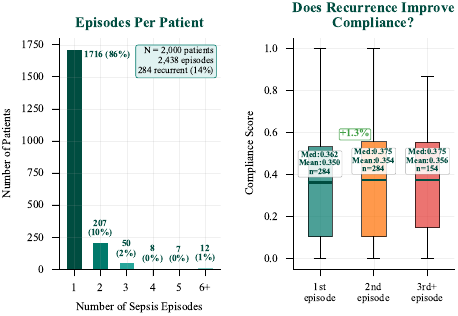}
    \caption{\footnotesize The left panel illustrates the distribution of sepsis episodes per patient across the study cohort ($N=2,000$ patients, 2,438 total episodes). While the majority of patients (86\%) experienced a single episode, 14\% presented with recurrent sepsis. The right panel evaluates whether clinical compliance improves in subsequent episodes for recurrent patients. A marginal improvement in median compliance is observed, increasing from 0.362 in the first episode to 0.375 in the second and third+ episodes, representing a +1.3\% gain in median performance between the first two clinical encounters.}
    \label{fig:patientvsepi}
\end{figure}

\section{Comparison Analysis}\label{app:companal}
\setlength{\tabcolsep}{4pt} 
\small

\begin{longtable}{p{0.16\textwidth} p{0.18\textwidth} p{0.18\textwidth} p{0.18\textwidth} p{0.18\textwidth}}

\caption{Comparison of SSC Bundle Compliance Gap Papers}
\label{tab:ssc-comparison} \\
\hline
\textbf{Parameter} & \textbf{\cite{gao2005impact}} & \textbf{\cite{damiani2015effect}} & \textbf{\cite{you2022relationship}} & \textbf{Our Work} \\
\hline
\endfirsthead

\multicolumn{5}{c}{{\bfseries Table \thetable\ continued from previous page}} \\
\hline
\textbf{Parameter} & \textbf{\cite{gao2005impact}} & \textbf{\cite{damiani2015effect}} & \textbf{\cite{you2022relationship}} & \textbf{Our Work} \\
\hline
\endhead

\hline
\multicolumn{5}{r}{{Continued on next page...}} \\
\endfoot

\hline
\endlastfoot

\textbf{Type of Study} 
& Prospective observational study 
& Systematic review \& meta-analysis 
& Retrospective observational cohort 
& SME guided computational pipeline \\
\hline

\textbf{SSC Bundle Gaps Identified} 
& Non-compliance with 6-hr and 24-hr bundles linked to higher hospital mortality 
& Suboptimal compliance across healthcare settings despite performance improvement programs 
& Bundle adherence varies significantly by ED admission time (off-hours worse) 
& Hour-1 bundle mean compliance of only 36.7\%; antibiotic timing (R2) worst at 0.24 mean fuzzy score \\
\hline

\textbf{Compliance Measurement Method} 
& Binary --- compliant vs.\ non-compliant per bundle window 
& Binary --- pooled compliance rates across 50+ studies 
& Binary --- adhered/not-adhered per bundle element 
& Graded fuzzy scores (0--1) per each of 8 SSC rules via Sugeno FIS \\
\hline

\textbf{Granularity of Gap Analysis} 
& Bundle-level (6-hr vs.\ 24-hr) 
& Program-level across institutions 
& Admission-time-level (day vs.\ night shift) 
& Per-rule, per-episode, per-phase (Hour-1, Resuscitation, Response) across 2,438 episodes \\
\hline

\textbf{Automated at Scale} 
& Manual prospective data collection 
& Manual aggregation of existing studies 
& Manual retrospective review 
& Partially automated on (need to have SME) \\
\hline

\textbf{Handles Unstructured EHR Text} 
& Not addressed 
& Not addressed 
& Not addressed 
& MedGemma resolves trade names, abbreviations, and microbiology ambiguity \\
\hline

\textbf{Key Bundle Gap Finding} 
& Full 6-hr compliance reduces mortality significantly 
& Compliance remains suboptimal even with improvement programs 
& Off-hours ED admission leads to worse bundle adherence 
& 51\% drop-off at elevated lactate threshold; antibiotic timing \& lactate measurement are primary failure points \\
\hline

\textbf{Links Compliance to Patient Outcomes} 
& Hospital mortality 
& Mortality reduction rates 
& Shock outcomes by admission time 
& ICU LOS --- low compliance yields median 5.1 days vs.\ high compliance 3.8 days \\
\hline

\textbf{Per-Patient Actionable Insights} 
& Population-level only 
& Aggregated across studies 
& Cohort-level only 
& Per-episode scores enabling individual and population-level quality improvement \\
\hline

\textbf{Domain Expert Validation} 
& None 
& None 
& None 
& SME consultation mid-pipeline for boundary setting, rule priority ordering, and edge case resolution \\

\end{longtable}

\clearpage

\section{Combined Results}\label{app:combres}
\begin{table*}[h]
\centering
\caption{Comprehensive Results from the Expert-Guided Neuro-Symbolic Pipeline for Sepsis Compliance Assessment. Data derived from 2,438 sepsis episodes across 2,000 patients from MIMIC-IV v3.1.}
\footnotesize
\setlength{\tabcolsep}{8pt} 
\begin{tabular}{lr | lr}
\toprule
\textbf{Metric} & \textbf{Value} & \textbf{Metric} & \textbf{Value} \\
\midrule
\multicolumn{2}{l}{\textit{STUDY COHORT}} & \multicolumn{2}{l}{\textit{MISSING DATA}} \\
Total Sepsis Patients & 2,000 & R2: Antibiotic Timing & 31.6\% \\
Total Sepsis Episodes & 2,438 & R3: Lactate Timing & 24.4\% \\
Recurrent Sepsis (14\%) & 284 & R1: Blood Culture & 8.8\% \\
Unique Drug Strings & 1,691 & R5: Fluids & 6.7\% \\
\midrule
\multicolumn{2}{l}{\textit{NORMALIZATION}} & \multicolumn{2}{l}{\textit{CONDITIONAL ACTIVATION}} \\
Drug Class. Agreement & 94.26\% & R4: Repeat Lactate & 912 (37\%) \\
Cohen's Kappa ($\kappa$) & 0.65 & R5: Fluid Resusc. & 1,052 (43\%) \\
Embedding Val. Rate & 98.59\% & R6: Vasopressors & 776 (32\%) \\
Agreed Antibiotics & 126 & R7: MAP Recovery & 686 (28\%) \\
Agreed Vasopressors & 7 & R8: Lac. Clearance & 826 (34\%) \\
\midrule
\multicolumn{2}{l}{\textit{COMPLIANCE RULE ($\mu$)}} & \multicolumn{2}{l}{\textit{ICU OUTCOMES}} \\
R1: Blood Culture & 0.36 & Low Compliance LOS & 5.1 days \\
R2: Antibiotics & 0.24 & Med. Compliance LOS & 4.1 days \\
R3: Lactate & 0.39 & High Compliance LOS & 3.8 days \\
R4: Repeat Lactate & 0.40 & Compliance--LOS $r$ & $-0.026$ \\
R5: Fluid Resusc. & 0.41 & Trend (Days/1\% Comp.) & $-0.006$ \\
R6: Vasopressors & 0.73 & Sample Size ($n$) & 2,362 \\
\midrule
\multicolumn{2}{l}{\textit{TIMING \& OUTCOMES}} & \multicolumn{2}{l}{\textit{WORST-RULE IMPACT}} \\
Meet 1-Hr Target & 13\% & R2 (Antibiotic) LOS & 7.0 days \\
30--60 min Median LOS & 2.95 d & R3 (Lactate) LOS & 5.0 days \\
120--180 min Median LOS & 4.34 d & R1 (Blood Cult.) LOS & 4.1 days \\
360+ min Median LOS & 4.74 d & R5 (Fluids) LOS & 2.9 days \\
\midrule
\multicolumn{2}{l}{\textit{CLINICAL CASCADE}} & \multicolumn{2}{l}{\textit{PHASE PERFORMANCE ($\mu$)}} \\
Blood Culture Obtained & 77\% & Phase 1 (Hour-1) & 0.00 \\
IV/IM Antibiotic Given & 68\% & Phase 2 (Resuscitation) & 0.67 \\
Lactate Measured & 76\% & Phase 3 (Response) & 1.00 \\
Fluids Administered & 36\% & \multicolumn{2}{l}{\textit{RECURRENT SEPSIS}} \\
Vasopressor Started & 37\% & 1st $\to$ 2nd $\mu$ Imp. & +1.3\% \\
\midrule
\multicolumn{2}{l}{\textit{CASCADE DROP-OFFS}} & \multicolumn{2}{l}{\textit{EPISODE COMPLIANCE ($\mu$)}} \\
BC $\to$ Antibiotic & $\downarrow$23\% & 1st Episode & 0.362 \\
Antibiotic $\to$ Lactate & $\downarrow$11\% & 2nd Episode & 0.375 \\
Lac. $\to$ Elev. Lac. & $\downarrow$51\% & 3rd+ Episodes & 0.375 \\
\bottomrule
\end{tabular}
\end{table*}

\section{Research Question Summary}\label{app:rq}
\begin{enumerate}
    \item \textbf{[RQ1] Is a neuro-symbolic approach necessary where purely symbolic or neural systems fall short?}

    The regex system verified 220 clinical drug strings against documented drug information, while MedGemma resolved the remaining lexical variation that a purely synonym-matching system would have missed; the fuzzy inference system then transformed these normalized inputs into graded compliance scores that a purely neural end-to-end model could not guarantee to align with SSC safety boundaries. The dual-path compliance comparison (Figure \ref{fig:regvsgem}) confirms that 80 episodes (3.3\%) produce divergent scores when classifiers operate independently, with 53 exhibiting differences exceeding 5\%, demonstrating that neither symbolic nor neural components alone produce reliable compliance assessment across all episodes.

    \item \textbf{[RQ2] Can a hybrid classifier pipeline achieve reliable semantic normalization of clinical data?}

    The hybrid pipeline achieves 94.26\% drug agreement, $\kappa = 0.65$ substantial agreement, and 98.59\% embedding confirmation rate, demonstrating reliable semantic normalization of unstructured $MIMIC-IV$ clinical data.

    \item \textbf{[RQ3] Can a fuzzy inference system generate graded SSC compliance scores that provide actionable clinical insights aligns with real-world clinical decision-making patterns and reflects subject matter expert (SME) perspectives on sepsis management?}

    The Sugeno fuzzy inference system reveals stark stratification across eight bundle rules, with antibiotic timing (R2, $\mu$ = 0.24) as the most critical compliance failure and MAP recovery (R7, $\mu$ = 0.97) as near-acceptable. The domain expert validated these findings by explaining that pre-ICU antibiotic administration accounts for low R2 scores, while high conditional-rule performance reflects survivorship bias rather than genuine compliance excellence. The graded scoring enabled nuanced phase-level analysis showing median compliance rising from 0.00 in the Hour-1 bundle to 1.00 in treatment response, insights that binary compliance systems could not surface.

    \item \textbf{[RQ4] What compliance patterns emerge across the sepsis cohort, and how do they correlate with patient outcomes?}

    The compliance cascade reveals a 51\% patient drop-off at the elevated lactate threshold, with antibiotic timing and lactate measurement as the most data-sparse Hour-1 interventions. Episodes with higher overall bundle compliance achieve median ICU stays of 3.8 days compared to 5.1 days for low-compliance episodes. The antibiotic timing dose-response relationship shows median ICU stays rising from 2.95 days at 30–60 minutes to 4.74 days beyond six hours, and worst-rule analysis localizes the highest ICU burden to antibiotic timing (R2, median 7.0 days) and lactate measurement (R3, median 5.0 days), establishing Hour-1 intervention failures as the primary driver of prolonged critical care.

\end{enumerate}

\end{document}